\documentclass[letterpaper, 10 pt, conference]{ieeeconf}  % Comment this line out if you need a4paper
\pdfoutput=1

\IEEEoverridecommandlockouts                              % This command is only needed if 
\usepackage{graphics} % for pdf, bitmapped graphics files
\usepackage{epsfig} % for postscript graphics files
\usepackage{mathptmx} % assumes new font selection scheme installed
\usepackage{times} % assumes new font selection scheme installed
\usepackage{amsmath} % assumes amsmath package installed
\usepackage{amssymb}  % assumes amsmath package installed

\DeclareMathAlphabet{\mathcal}{OMS}{cmsy}{m}{n} % keep the mathptmx package, but with the original \mathcal symbols

\PassOptionsToPackage{hyphens}{url}
\usepackage[hyphens]{url}

\usepackage{algorithm}
\usepackage{algorithmic}

\usepackage{amsfonts} % overhead bar

\usepackage{subcaption}

\graphicspath{{figures/}} %Setting the graphicspath

\usepackage{balance} % equalize two columns of the last page

\title{\LARGE \bf
Optimal Reduced-order Modeling of Bipedal Locomotion
}

\author{Yu-Ming Chen$^{1}$ and Michael Posa$^{1}$% <-this % stops a space
\thanks{$^{1}$Yu-Ming Chen and Michael Posa are with the General Robotics, Automation, Sensing and Perception (GRASP) Laboratory, University of Pennsylvania, Philadelphia, PA 19104, USA
        {\tt\small \{yminchen, posa\}@seas.upenn.edu}}%
}

\begin{document}

\maketitle
\thispagestyle{empty}
\pagestyle{empty}

%%%%%%%%%%%%%%%%%%%%%%%%%%%%%%%%%%%%%%%%%%%%%%%%%%%%%%%%%%%%%%%%%%%%%%%%%%%%%%%%
\begin{abstract}
State-of-the-art approaches to legged locomotion are widely dependent on the use of models like the linear inverted pendulum (LIP) and the spring-loaded inverted pendulum (SLIP), popular because their simplicity enables a wide array of tools for planning, control, and analysis. 
However, they inevitably limit the ability to execute complex tasks or agile maneuvers. 
In this work, we aim to automatically synthesize models that remain low-dimensional but retain the capabilities of the high-dimensional system.
For example, if one were to restore a small degree of complexity to LIP, SLIP, or a similar model, our approach discovers the form of that additional complexity which optimizes performance. 
In this paper, we define a class of reduced-order models and provide an algorithm for optimization within this class.
To demonstrate our method, we optimize models for walking at a range of speeds and ground inclines, for both a five-link model and the Cassie bipedal robot.

\end{abstract}

%%%%%%%%%%%%%%%%%%%%%%%%%%%%%%%%%%%%%%%%%%%%%%%%%%%%%%%%%%%%%%%%%%%%%%%%%%%%%%%%
\section{Introduction}

\begin{figure}[t!]
 \centering
 \includegraphics[width=0.8\linewidth]{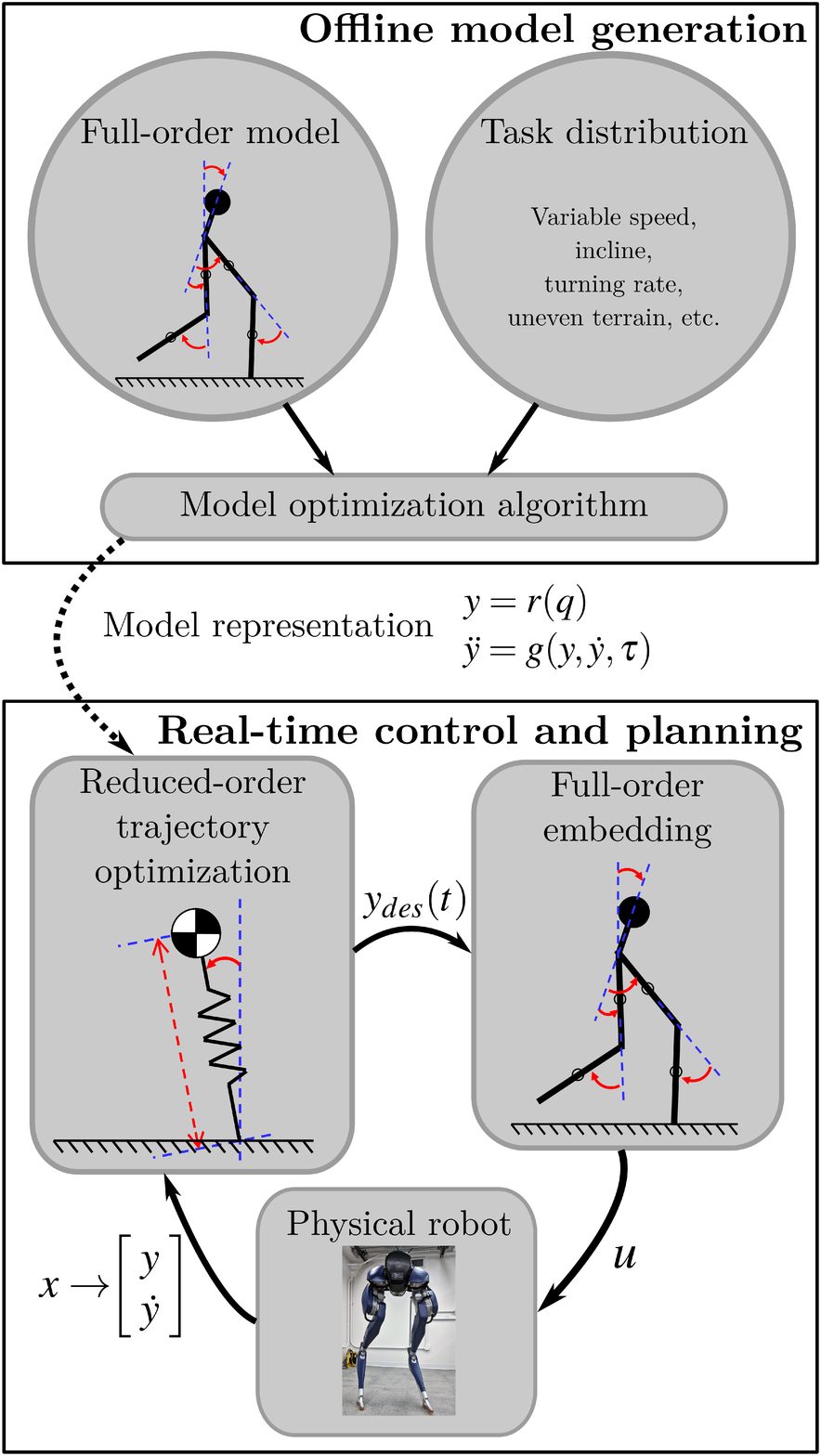}
 \caption{An outline of the synthesis and deployment of optimal reduced-order models. Offline, given a full-order model and a distribution of tasks, we optimize a new model that is effective over the task space. Online, we generate new plans for the reduced-order model and track these trajectories on the true, full-order system.}
 \label{fig:outline}
\end{figure}

Modern legged robots, like the Agility Robotics Cassie, have many degrees of freedom, tens or more actuators and may have passive dynamic elements such as springs and dampers.
To manage this complexity, and simplify the process of planning and control design, the community has embraced the use of reduced-order models.
Particularly popular are the linear inverted pendulum (LIP) \cite{Kajita91, Kajita01}, the spring-loaded inverted pendulum (SLIP) \cite{blickhan1989spring}, and various permutations.
The LIP has a long history as a predominant approach in robot walking, and formed the basis of many approaches taken during the DARPA robotics challenge \cite{Feng15, Koolen_etal2013, Kuindersma14}. 
The SLIP is widely used to explain energy efficient running \cite{hutter2010slip, poulakakis2009spring, hubicki2016walking, martin2017experimental}. 
These models have been empirically shown to capture the dominant dynamics of the robots in particular tasks, and their simplicity enables solutions to the challenging problems of control and planning design. 
For example, many of locomotion planning problems can be solved in realtime with the low-dimensional models \cite{apgar2018fast}. 

The downside, however, is that by forcing robots to act like a low-degree-of-freedom model, these approaches restrict the motion of complex robots and necessarily sacrifices performance.
This can result in energetically inefficient motion, or fail to extend to wide range of tasks. 
For example, the LIP greatly restricts both efficiency and stride length.
%Furthermore, any model with a massless swing foot would fail to describe the impact dynamics.
% Michael--removed that line since it is not true.
These limitations have long been acknowledged by the community, resulting in a wide array of extensions that universally rely on human intuition, and are generally in the form of mechanical components (a spring, a damper, a joint, a rigid body with inertia, etc) \cite{pratt1997virtual, sellaouti2006faster, hutter2011scarleth, garofalo2012walking, faraji20173lp, koolen2012capturability, libby2016comparative}. 
The ad hoc nature of these extensions demonstrates that the community implicitly admits both that the simplest models are insufficient, and that it is not known which extensions are most beneficial. 
Additionally, it has been shown that not all model extensions improve the performance of robots much. 
For example, allowing center of mass height to vary provides limited aid in the task of balancing \cite{Posa17, Koolen16}.

The primary contribution of this paper is an optimization algorithm to automatically synthesize new reduced order models, embedding high-performance capabilities within low-dimensional representations.
Given a distribution of tasks, and a nominal full-order model, we propose a bilevel optimization of stochastic gradient descent and trajectory optimization to search within a broad class of simple models.

%%%
% How to bring up the downside of complex model planning?

\section{Background}\label{sec:background}

\begin{figure}[t]
 \centering
 \includegraphics[width=1\linewidth]{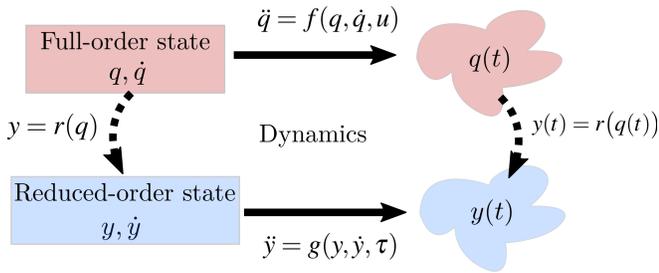}
 \caption{Relationship of the full-order and reduced-order models. The generalized positions $q$ and $y$ satisfy the embedding function $r$ for all time, and the evolution of the velocities $\dot{q}$ and $\dot{y}$ respects the dynamics $f$ and $g$, respectively.}
 \label{fig:mapping}
\end{figure}

\subsection{Walking via simple models}
A good simple model is a low-dimensional representation  that captures much of the relevant dynamics while enabling effective control design, a concept closely related to that of templates and anchors \cite{Full1999, Saranli2001}.
One observation, common to many approaches, lies in the relationship between foot placement, ground reaction forces, and the center of mass (COM).
While focusing on the COM neglects the individual robot limbs, controlling the COM position has proven to be an excellent proxy for the stability of a walking robot.
COM-based simple models include the LIP \cite{Kajita91,Kajita01}, which restricts angular momentum and vertical motion, SLIP \cite{blickhan1989spring}, hopping models \cite{Raibert1984}, inverted pendulums \cite{Garcia1998,Schwab2001,Gregg12,Bhounsule2014, Hereid2014}, and others.
Since these models are universally low-dimensional, they have enabled a variety of control synthesis and analysis techniques that would not otherwise be computationally tractable.
For example, numerical methods have been successful at finding robust gaits and control designs \cite{Byl2009,OguzSaglam2014, Kelly2015, Koolen12}, and assessing stability \cite{Pratt06}.
A common approach, which we also adopt in this work, is to first plan motions of the reduced-order model, and then track this lower-dimensional trajectory with a technique like operational space control \cite{Wensing2013}.

Despite their successes, many deficiencies have been found in these simplest models (e..g \cite{Suzuki2006}).
For example, by eliminating the use of angular momentum and prohibiting impacts, the LIP greatly reduces energy efficiency and limits speed and stride length.
This has necessitated  extensions (e.g. \cite{faraji20173lp} and others), where all replace inverted pendulum abstractions with that of a more complex physical model.

\subsection{Trajectory optimization}
This paper will heavily leverage trajectory optimization within the inner loop of a bilevel optimization problem.
We briefly review the area here, but the reader is encouraged to see \cite{Betts01} for a more complete description.
Generally speaking, trajectory optimization is a process of finding state $x(t)$ and input $u(t)$ that minimize some measure of cost $h$ while satisfying a set of constraints $C$.
Following the approach taken in prior work \cite{Posa13, Posa2016}, we explicitly optimize over state, input, and constraint (contact) forces $\lambda(t)$,

\begin{equation}\label{eq:trajopt}
    \begin{array}{cl}
     \underset{x(t),u(t),\lambda(t)}{\text{min}} & \displaystyle\int_{t_0}^{t_f} h(x(t),u(t)) dt  \\
     \text{s.t.} & \dot{x}(t) = f(x(t),u(t),\lambda(t)), \\
      & C (x(t),u(t),\lambda(t)) \leq 0,  \\
    \end{array}
\end{equation}
where $f$ is the dynamics of the system, $\lambda$ are the forces required to satisfy holonomic constraints, and $t_0$ and $t_f$ are the initial and the final time respectively.
Standard approaches  discretize in time, formulating \eqref{eq:trajopt} as a finite-dimensional nonlinear programming problem.
For the purposes of this paper, any such method would be appropriate;  we use DIRCON \cite{Posa2016} to address the closed kinematic chains present in the Cassie robot.
DIRCON transcribes the infinite dimensional problem in (\ref{eq:trajopt}) into a finite dimensional nonlinear problem 
\begin{equation}\label{eq:disc_trajopt}
    \begin{array}{cl}
     \underset{w}{\text{min}} & \displaystyle\sum_{i=1}^{n-1} \frac{1}{2} \big( h(x_i,u_i) + h(x_{i+1},u_{i+1}) \big) \delta_k  \\
     \text{s.t.} & f_c(x_i,x_{i+1},u_i,u_{i+1},\lambda_i,\lambda_{i+1},\delta_i,\alpha_i)=0, \\
      & \hspace{42mm} i=1,...,n-1\\
      & C (x_i,u_i,\lambda_i) \leq 0, \hspace{19mm} i=1,...,n\\
    \end{array}
\end{equation}
where $n$ is the number of knot points, $f_c$ is the collocation constraint for dynamics, and the decision variables are
\begin{equation*}
\begin{alignedat}{2}
  \hspace{5mm} w = [x_1,...,x_n,&u_1,...,u_n, \lambda_1,...,\lambda_n, \\
  &\delta_1, ..., \delta_{n-1}, \alpha_1, ..., \alpha_{n-1}]^\top \in \mathbb{R}^{n_w}.
\end{alignedat}
\end{equation*}
where $\delta_i$'s are time intervals, and $\alpha_i$'s are slack variables specific to DIRCON.

\section{Approach}\label{sec:approaches}

In this section, we first propose a concrete definition of reduced-order models, along with a notion of quality (or cost) for such models.
We then introduce a bilevel optimization algorithm to optimize within our class of models.

\subsection{Definition of reduced-order models}
\label{sec:definition}
Let $q$ and $u$ be the generalized position and input of the full-order model, and let $y$ and $\tau$ be the generalized position and input of the reduced-order model.
We define a reduced-order model $\mu$ of dimension $n_y$ by two things -- an embedding function $r: q \mapsto y(q)$ and the second-order dynamics of the reduced-order model $g(y, \dot y, \tau)$. 
\begin{equation}\label{eq:model_def}
\begin{aligned}
\mu & \triangleq (r,g),
\end{aligned}
\end{equation}
with 
\begin{equation}\label{eq:model}
\begin{aligned}
y&=r(q),  \\
\ddot{y}&= g(y,\dot{y},\tau),
\end{aligned}
\end{equation}
where $\dim y < \dim q$ and $\dim \tau \leq \dim u$. 
As an example, to represent SLIP, $r$ is the spring length and the spring angle with respect to the normal direction of ground, $g$ is the spring-mass dynamics, and $\dim \tau = 0$ as SLIP is passive.

Fig. \ref{fig:mapping} shows the relationship between the two models. 
If we integrate the two models forward in time with their own dynamics, the resultant trajectories will still satisfy the embedding function $r$ at any time in the future. 

\subsection{Problem statement}
As shown in the upper half of Fig. \ref{fig:outline}, the goal is to find an optimal model $\mu^*$, given a distribution over a set of tasks $\Gamma$. 
The distribution could be provided \textit{a priori} or estimated via the output of a higher-level motion planner.
The tasks in $\Gamma$ might include anything physically achievable by the robot, such as walking up a ramp at different speeds, turning at various rates, jumping, running with least amount of energy, etc.
The goal, then, is to find a reduced-order model that enables low-cost motion over the space of tasks,
\begin{equation}\label{eq:model_opt}
    \mu^* = \underset{\mu \in M}{\text{argmin}} \  \mathbb{E}_{\gamma} \left[ \mathcal{J}_{\gamma}(\mu) \right], 
\end{equation}
where $M$ is the model space, $\mathbb{E}_\gamma$ takes the expected value over $\Gamma$, and $\mathcal{J}_{\gamma}(\mu)$ is the cost required to achieve a task $\gamma \in \Gamma$ while the robot is restricted to a particular model $\mu$. 

With our definition of model in (\ref{eq:model_def}), this is an infinite dimensional problem over the space of embedding and dynamics functions, $r$ and $g$.
To simplify, we express  $r$ and  $g$ in terms of specified feature functions $\{\phi_{e,i} \mid i = 1, \ldots, n_e\}$ and $\{\phi_{d,i} \mid i = 1, \ldots, n_d\}$ with linear weights $\theta_e \in \mathbb{R}^{n_y \cdot n_e}$ and $\theta_d \in \mathbb{R}^{n_y \cdot n_d}$.
Further assuming that the dynamics are affine in $\tau$ with constant multiplier,  $r$ and $g$ are given as
\begin{subequations}\label{eq:parametrized_rom}
\begin{align}
  y&=r(q;\theta_e) \hspace{6.2mm} = \Theta_e \phi_e (q), \label{eq:parametrized_rom_1}\\
  \ddot{y}&= g(y,\dot{y},\tau; \theta_d)=  \Theta_{d} \phi_{d} (y,\dot{y}) + B\tau, \label{eq:parametrized_rom_2}
\end{align}
\end{subequations}
where $\Theta_e \in \mathbb{R}^{n_y \times n_e}$ and $\Theta_d \in \mathbb{R}^{n_y \times n_d}$ are $\theta_e$ and $\theta_d$ arranged as matrices. $\phi_e = [\phi_{e,1},\ldots,\phi_{e,n_e}]^\top$, $\phi_d = [\phi_{d,1},\ldots,\phi_{d,n_d}]^\top$, and $B \in \mathbb{R}^{n_y \times n_\tau}$.
While we choose linear basis functions, note that any differentiable function approximator (e.g. a neural network) might be equivalently used.

We can see $\theta = [\theta_e^\top, \theta_d^\top]^\top \in \mathbb{R}^{n_t}$ parameterize the model class $M$.
Therefore, we can rewrite (\ref{eq:model_opt}) as  
\begin{equation}\label{eq:parametrized_model_opt}
    \theta^* = \underset{\theta}{\text{argmin}} \  \mathbb{E}_{\gamma} \left[ \mathcal{J}_{\gamma}(\theta) \right]. 
    \tag{O}
\end{equation}
From now on, we work explicitly in $\theta$, rather than $\mu$.

\subsection{Task evaluation}
We use trajectory optimization to evaluate the task cost $\mathcal{J}_{\gamma}(\theta)$.
Under this setting, a task $\gamma$ is defined by a cost $h_\gamma$ and task-specific constraints $C\gamma$. 
$\mathcal{J}_{\gamma}(\theta)$ is the cost to achieve task $\gamma$ while simultaneously respecting the embedding and dynamics given by $\theta$.
The resulting optimization problem is similar to (\ref{eq:disc_trajopt}), but contains additional constraints and decision variables for the reduced-order model embedding,
\begin{equation}\label{eq:model_trajopt}
\begin{array}{rl}
    \hspace{-2mm} \mathcal{J}_{\gamma}(\theta) \triangleq \ \underset{w}{\text{min}} \hspace{-2mm} & \displaystyle\sum_{i=1}^{n-1} \frac{1}{2} \big( h_\gamma(x_i,u_i,\tau_i) + h_\gamma(x_{i+1},u_{i+1},\tau_{i+1}) \big) \delta_k  \\
      \text{s.t.} \hspace{-1.5mm} & f_c(x_i,x_{i+1},u_i,u_{i+1},\lambda_i,\lambda_{i+1},\delta_i)=0, \\
      & \hspace{43.7mm} i=1,\ldots,n-1 \\
      & g_c\left(x_i, x_{i+1}, \tau_i, \tau_{i+1}, \delta_i;\theta\right) = 0, \hspace{1.9mm} i=1,\ldots,n-1 \\
      & C_\gamma (x_i,u_i,\lambda_i) \leq 0, \hspace{19.3mm} i=1,\ldots,n \\
\end{array}
\tag{TO}
\end{equation}
where $f_c$ and $g_c$ are collocation constraints for the full-order and reduced-order dynamics. 
The decision variables are $w~=~[x_1,...,x_n,$ $u_1,...,u_n,$ $ \lambda_1,...,\lambda_n,$ $ \tau_1,...,\tau_n,$ $\delta_1, ..., \delta_{n-1},$ $\alpha_1, ..., \alpha_{n-1}]^\top,$
noting the addition of $\tau_i$.
%\begin{equation}\label{eq:model_trajopt}
%    \begin{array}{cl}
%     \underset{x(t),u(t),y(t),\tau(t)}{\text{min}} & \displaystyle\int_{t_0}^{t_f} h_\gamma(x(t),u(t),\tau(t)) dt  \\
%     \text{s.t.} & \dot{x}(t) = f(x(t),u(t)) \\
%      & y(t) = r(q(t);\theta_1) \\
%      & \ddot{y}(t) = g\left(y(t), \dot{y}(t), \tau(t);\theta_2\right) \\
%      & C_\gamma (x(t),u(t)) \leq 0  \\
%    \end{array}
%    \tag{TO}
%\end{equation}
Observe that this problem is equivalent to simultaneous optimization of full-order and reduced-order trajectories that must also be consistent with the embedding $r$.
We solve the nonlinear problem in \eqref{eq:model_trajopt} using the SNOPT toolbox \cite{Gill05} .

The formulation of dynamics and holonomic constraints of the full-order model are described in \cite{Posa2016}, so here we only present $g_c$.
We approximate each segment of the trajectory $y$ by a cubic polynomial $y_p(t)$. 
\begin{equation*}
y_p(t) = a_0 + a_1 t + a_2 t^2 + a_3 t^3,
\end{equation*}
where $a_0$ to $a_3$ are coefficients and can be solved for in terms of $x_i$ and $x_{i+1}$ when we impose boundary conditions
\begin{equation*}
\begin{alignedat}{2}
y_p(0) & = y_i & = &  r (q_i;\theta_e), \\
\dot{y}_p(0) & = \dot{y}_i & = & \frac{\partial r (q_i;\theta_e)}{\partial q_i} \dot{q}_i,\\
y_p(\delta_i) & = y_{i+1} & = & r (q_{i+1};\theta_e), \\ 
\dot{y}_p(\delta_i) & = \dot{y}_{i+1} & = & \frac{\partial r (q_{i+1};\theta_e)}{\partial q_{i+1}} \dot{q}_{i+1}.
\end{alignedat}
\end{equation*}
In the line of standard direct collocation \cite{Hargraves87}, $g_c = 0$ requires that the second time-derivative of $y_p$ must match the dynamics in (\ref{eq:parametrized_rom_2}) at the collocation point. That is, 
\begin{equation}\label{eq:rom_dynamics}
\begin{alignedat}{2}
&\ \ddot{y}_p(\tfrac{\delta_i}{2}) =  g(y_c,\dot{y}_c,\tau_c; \theta_d)\Rightarrow \frac{\dot{y}_{i+1} - \dot{y}_i}{\delta_i} \ & - & \  g(y_c,\dot{y}_c,\tau_c; \theta_d) = 0\\
\end{alignedat}
\end{equation}
%\begin{equation}\label{eq:rom_dynamics}
%\begin{alignedat}{2}
%\Rightarrow \ &  \frac{\dot{y}_{i+1} + \dot{y}_i}{2} \ & - & \  \Theta_{d} \phi_{d} (y_c,\dot{y}_c) + B\tau_c = 0,
%\end{alignedat}
%\end{equation}
where 
\begin{equation*}
\begin{alignedat}{2}
y_c \ & = y_p(\frac{\delta_i}{2})  =  \frac{1}{2}(y_i + y_{i+1}) + \frac{\delta_i}{8}(\dot{y}_i - \dot{y}_{i+1}), \\
\dot{y}_c \ & = \dot{y}_p(\frac{\delta_i}{2})  =  \frac{3}{2 \delta_i}(- y_i + y_{i+1}) - \frac{1}{4}(\dot{y}_i + \dot{y}_{i+1}) ,\\
\tau_c \ & =  \frac{\tau_i + \tau_{i+1}}{2}.
\end{alignedat}
\end{equation*}

\begin{algorithm}[t!]
\caption{Reduced-order model optimization}
\begin{algorithmic}[1]\label{alg:model_opt}
\renewcommand{\algorithmicrequire}{\textbf{Input:}}
\renewcommand{\algorithmicensure}{\textbf{Output:}}
\REQUIRE  $\Gamma$
\ENSURE  $\theta^*$
\\ \textit{Model initialization}
\STATE $\theta \leftarrow \theta_0$
\\ \textit{Model optimization}
\REPEAT
\STATE Randomly sample tasks $\gamma_j \sim \Gamma$ for $j=1,\ldots,N.$
\FOR {$j=1,\ldots,N$}
\STATE Solve (\ref{eq:model_trajopt}) to get $\mathcal{J}_{\gamma_j}(\theta)$
\STATE Calculate $\nabla_\theta \left[\mathcal{J}_{\gamma_j}(\theta) \right]$
\ENDFOR
\STATE Average the gradients $\Delta \theta = \frac{\sum_{j=1}^{N} \nabla_\theta \left[ \mathcal{J}_{\gamma_j}(\theta) \right]}{N}$
\STATE \textit{Gradient descent} $\theta \leftarrow \theta - d\cdot \Delta \theta$
\UNTIL convergence
\RETURN $\theta$
\end{algorithmic}
\end{algorithm}

\subsection{Bilevel optimization algorithm}

Since there might be a large or infinite number of tasks $\gamma \in \Gamma$ in (\ref{eq:parametrized_model_opt}), solving for the exact solution is often intractable. 
Therefore, we use stochastic gradient descent as the outer loop (to trajectory optimization in the inner loop).
That is, we sample a set of tasks from the distribution of $\Gamma$ and optimize the averaged sample cost over the model parameters $\theta$. 

The full approach to (\ref{eq:parametrized_model_opt}) is outlined in Algorithm \ref{alg:model_opt}. 
Starting from an initial parameter seed $\theta_0$, $N$  tasks are sampled, and the cost for each task is evaluated by solving the corresponding trajectory optimization problem (\ref{eq:model_trajopt}). 

To compute each gradient  $\nabla_\theta \left[ \mathcal{J}_{\gamma_j}(\theta) \right]$, we adopt an approach based in sequential quadratic programming. We locally approximate (\ref{eq:model_trajopt}) with an equality-constrained quadratic program, only considering the active constraints. 
Let $\Tilde{w}_\gamma = w - w^*_\gamma$ and $\Tilde{\theta} = \theta - \theta^{(i)}$, where $w^*_\gamma$ is the optimal solution of (\ref{eq:model_trajopt}), and $\theta^{(i)}$ is the parameter at the $i$-th iteration. 
The approximated quadratic program is 
\begin{equation}\label{eq:QP}
    \begin{array}{cl}
     \mathcal{J}_{\gamma}(\theta) \  \approx \  \underset{w_\gamma}{\text{min}} & \frac{1}{2}\Tilde{w}^T_\gamma H_\gamma \Tilde{w}_\gamma + b^T_\gamma\Tilde{w}_\gamma + c_\gamma \hspace{5mm}\\
      \hspace{16mm} \text{s.t.} & F_\gamma\Tilde{w}_\gamma+G_\gamma\Tilde{\theta}=0 \\ 
    \end{array}
\end{equation}
Using the KKT conditions, we can derive the following equation for the optimal solution
\begin{equation}\label{eq:KKT}
\left[\begin{array}{cl}
    H_\gamma & F^T_\gamma\\
    F_\gamma & 0\\
    \end{array}\right]
\left[\begin{array}{c}
    \Tilde{w}^*_\gamma \\
    \nu^*_\gamma \\
    \end{array}\right] = 
\left[\begin{array}{c}
    -b_\gamma \\
    -G_\gamma\Tilde{\theta} \\
    \end{array}\right],
\end{equation}
where $\nu^*_\gamma$ is the optimal dual solution.
(\ref{eq:KKT}) can be further solved, with the solution rewritten as
\begin{equation}\label{eq:w_star}
\Tilde{w}^*_\gamma = Q_\gamma\Tilde{\theta} + p_\gamma,
\end{equation}
for some $Q_\gamma \in \mathbb{R}^{n_w \times n_t}$ and $p_\gamma \in \mathbb{R}^{n_w}$. 
Since we approximate the original problem around $w^*$ and $\theta^{(i)}$, we know that $\tilde{w}^*_\gamma = 0$ if $\tilde{\theta} = 0$; therefore, $p_\gamma = 0$. 
Substituting \eqref{eq:w_star} into \eqref{eq:QP} and taking the gradient, we derive
\begin{equation}\label{eq:gradient_J}
\nabla_\theta \left[\mathcal{J}_{\gamma_j}(\theta) \right]\Big\rvert_{\theta=\theta^{(i)}} = Q_{\gamma_j}^\top b_{\gamma_j}.
\end{equation}

The algorithm is deemed to have converged if the norm of the gradient falls below a specified threshold.

\section{Examples}\label{sec:examples}

\begin{figure}[t!]
 \centering
 \includegraphics[width=.8\linewidth]{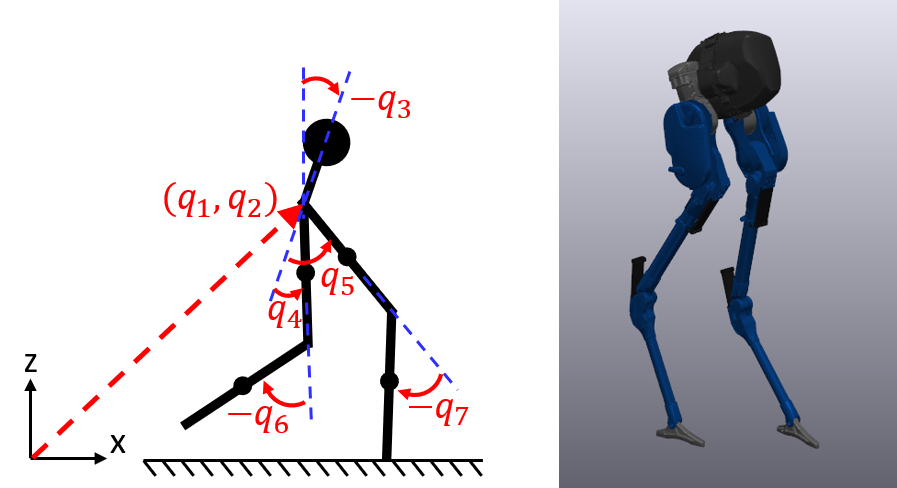}
 \caption{Examples of full-order models. On the left is a five-link robot. On the right is the 3D Cassie biped, which has five actuators per leg.}
 \label{fig:cassieandplanarrobot}
\end{figure}

\begin{figure}[t!]
 \centering
 \includegraphics[width=0.8\linewidth]{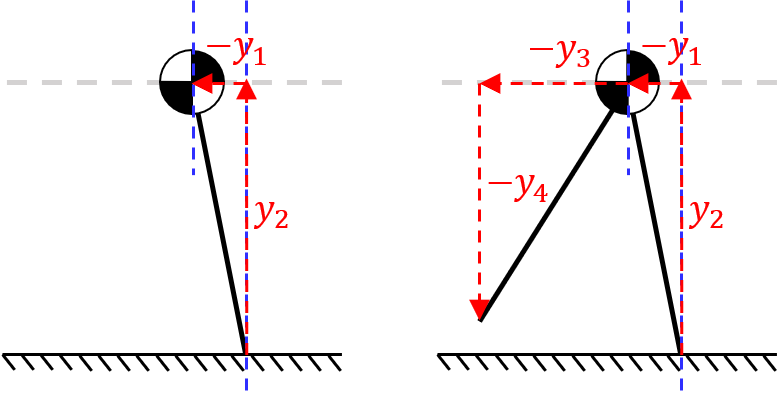}
 \caption{Two initial reduced-order models with generalized positions. On the left is the standard LIP; on the right is an LIP with an actuated swing foot.}
 \label{fig:initialrom}
\end{figure}

\begin{figure*}[t!]
\centering
  \begin{subfigure}[t]{.4\textwidth}
    \centering
    \includegraphics[width=\linewidth]{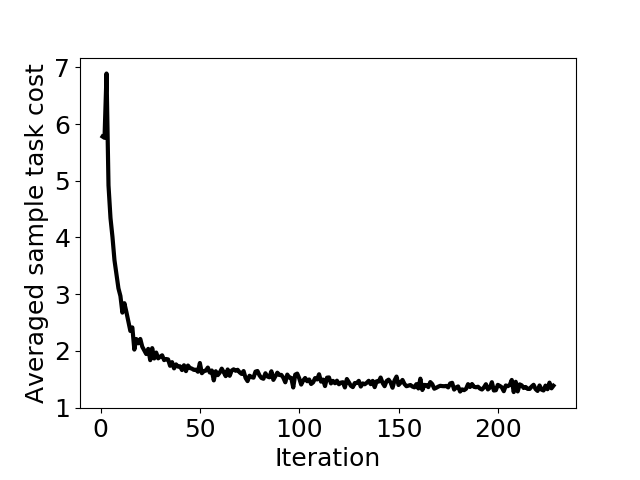}
	\caption{2D reduced-order model embedded in the five-link planar robot.}
	\label{fig:planarrobot_2drom_cost}
  \end{subfigure}
  \hspace{10mm}%
  \begin{subfigure}[t]{.4\textwidth}
    \centering
    \includegraphics[width=\linewidth]{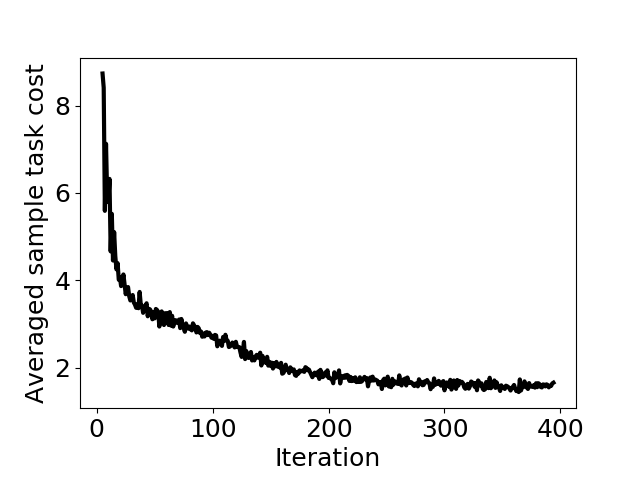}
	\caption{4D reduced-order model embedded in the five-link planar robot.}
	\label{fig:planarrobot_4drom_cost}
  \end{subfigure}

  \smallskip

  \begin{subfigure}[t]{.4\textwidth}
    \centering
    \includegraphics[width=\linewidth]{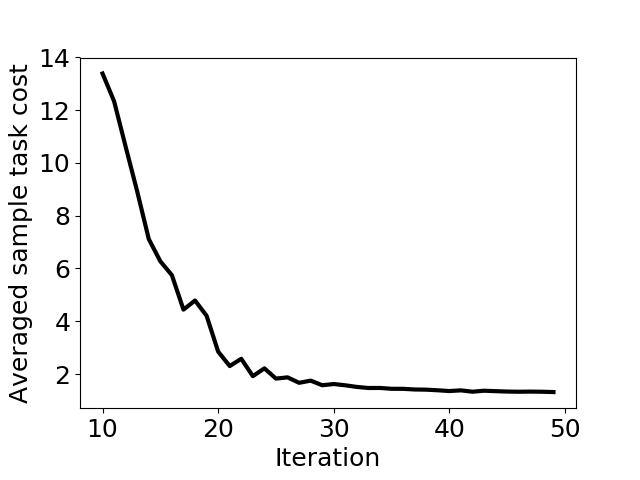}
	\caption{2D reduced-order model embedded in Cassie.}
	\label{fig:cassie_2drom_cost}
  \end{subfigure}
  \hspace{10mm}%
  \begin{subfigure}[t]{.4\textwidth}
    \centering
    \includegraphics[width=\linewidth]{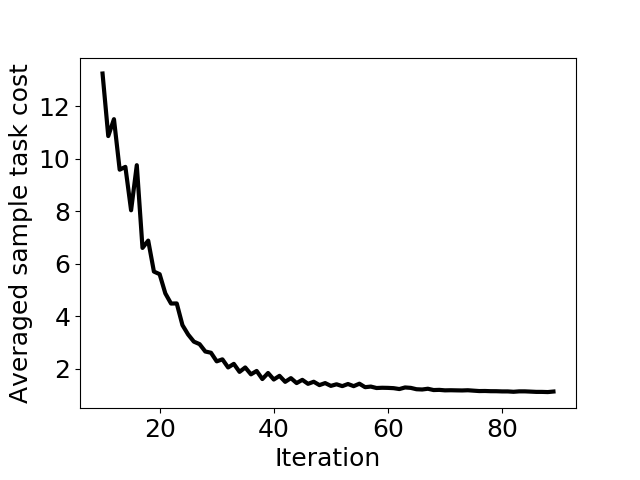}
	\caption{4D reduced-order model embedded in Cassie.}
	\label{fig:cassie_4drom_cost}
  \end{subfigure}
 \caption{The averaged cost of the sampled tasks over iterations. Costs are normalized by the cost associated with the full-order model (a lower bound on any reduced-order model).}
 \label{fig:cost}
\end{figure*}

We test our algorithm of model optimization with two robots: a five-link planar robot and the three dimensional Cassie biped (Fig. \ref{fig:cassieandplanarrobot}).
For all examples, the motions of the optimized models are shown in the accompanying video.
Examples were generated using the Drake software toolbox \cite{Drake2016} and source code is freely available\footnote{https://github.com/DAIRLab/dairlib/tree/goldilocks-model-dev}.

The planar robot consists of five links with non-zero mass and inertia, and four actuated joints with torque saturation. 
The thighs and shanks are of mass 2.5 kg and length 0.5 m. 
The weight of the torso is 10 kg and the length is 0.3 m. 
The robot has point feet, and the contacts between feet and ground are perfectly inelastic. 

The Cassie has stiff rotational springs in the knees and ankles.
Here, however, we simplify the model by treating these springs as infinitely stiff; this simplification is necessary for the coarse integration steps used in trajectory optimization, and has been used successfully with Cassie \cite{hereid2018rapid}.
The legs also contain two four-bar linkages, which we model with a fixed-distance constraint and corresponding constraint force.
There are five motors on each leg. Three located at the hip, one at the knee, and one at the toe. 

The hybrid equations of motion of either robot are written
\begin{equation}\label{eq:EoM}
    \begin{cases}
    
\begin{alignedat}{2}
\dot{x} &= f(x, u),& \hspace{5mm} &x^- \not\in S \\
x^+ &= \Delta(x^-,\Lambda),& &x^- \in S
\end{alignedat}
    
    \end{cases}
\end{equation}
where $x^-$ and $x^+$ are pre- and post-impact state, $\Lambda$ is the impulse of swing foot touchdown,
$\Delta$ is the discrete mapping of the touchdown event, and $S$ is the surface in the state space where the event must occur. 

We assume the robot walks with instantaneous change of support. 
That is, the robot transitions from right support to left support instantaneously, and vice versa. 
Therefore, the phase sequence cycles through a single support phase. 
For the examples here, we consider only half-gait periodic motion, and so include right-left leg alternation in the impact map $\Delta$.

\subsection{Initial reduced-order models}
To demonstrate the algorithm, we optimize two models for each robot.
The first is 2 dimensional, with 0 inputs, and the second is 4 dimensional with 2 inputs.
We initialize the 2D model with an LIP and the 4D model with an LIP plus a point-mass swing foot (Fig. \ref{fig:initialrom}). 
The generalized positions $y$ for both models are shown in Fig. \ref{fig:initialrom}. 
For reference, the equations of motion of the LIP with a point-mass swing foot are 
\begin{equation}\label{eq:lipmwithsfdyn}
\ddot{y} = \left[\begin{array}{c}
     \ddot{y}_1\\
     \ddot{y}_2\\
     \ddot{y}_3\\
     \ddot{y}_4\\
    \end{array}\right] =
    \left[\begin{array}{c}
     c_g \cdot y_1 / y_2\\
     0\\
     0\\
     0\\
    \end{array}\right] + 
    \left[\begin{array}{cc}
     0& 0\\
     0& 0\\
     1& 0\\
     0& 1\\
    \end{array}\right] \left[\begin{array}{c}
     \tau_1\\
     \tau_2\\
    \end{array}\right],
\end{equation}
where $c_g=9.81~m/s^2$ is the  gravitational acceleration.
For the LIP, the dynamics are given in the first two rows of \eqref{eq:lipmwithsfdyn}.

%\begin{equation}\label{eq:lipmdyn}
%\ddot{y} = \left[\begin{array}{c}
%     \ddot{y}_1\\
%     \ddot{y}_2\\
%    \end{array}\right] =
%    \left[\begin{array}{c}
%     c_g \cdot y_1 / y_2\\
%     0\\
%    \end{array}\right],
%\end{equation}

\subsection{Five-link planar robot}
For the case of five-link robot, we have $q\in \mathbb{R}^7$, where the first 3 elements are the floating-base joint. Recall that the contact constraint with the ground is solved implicitly.

We choose a rich feature set $\phi_e$ that includes the COM position with respect to the stance foot, the swing foot position with respect to the center of mass, the hip position $(q_1, q_2)$, and all quadratic combinations of the elements in $\{1, \text{cos}(q_3), \text{sin}(q_3), ..., \text{cos}(q_7), \text{sin}(q_7)\}$.
% i.e. $\{a_i a_j \mid a_i, a_j \in A\}$.

For the 2D model, the feature set $\phi_d$ includes $c_g \cdot y_1 / y_2$, and all quadratic combinations of the elements in $\{1, y_1, y_2, \dot{y}_1, \dot{y}_2\}$. 
%For the 4D model, the feature set $\phi_2$ includes $c_g \cdot y_1 / y_2$, and all quadratic combinations of the elements in $\{1, y_1, ..., y_4, \dot{y}_1, ..., \dot{y}_4\}$.
For the 4D model, the feature set $\phi_2$ is constructed in a similar way.
Note that these feature vectors were chosen to explicitly include elements of the LIP and the LIP with a swing foot, but also include a diverse set of additional terms.
Initial parameters $\theta$ can be easily chosen to match the LIP-based initial models.

We chose $\Gamma$ to include walking with different speeds between $0.27$ and $0.54$ m/s and on ground inclines between $-0.08$ and $0.08$ radians. 
The cost $h_\gamma$ is the sum of weighted norm of generalized velocity $\dot{q}$, input of the robot $u$ and input of the reduced-order model $\tau$. 
We include $\tau$ in the cost to regularize the input to the reduced-order model, and to correlate it with cost on the original model.
% Michael: deleted these lines. If they are to be included, they belong in the conclusion.
%Currently there is not much experiment done toward analyzing the benefit of $\tau$ in cost function. 
%The comparison of different cost functions is a part of the future work.

Fig. \ref{fig:planarrobot_2drom_cost} and \ref{fig:planarrobot_4drom_cost} show results from optimization, where the empirical average cost decreases rapidly during the process.
Note that the empirical average does not strictly decrease, as tasks are randomly sampled and are of varying difficulty.
The optimized model is capable of expressing lower cost and more natural motion than the LIP, better leveraging the natural dynamics of the five-link model.

\subsection{Cassie}

For the 3D Cassie, the generalized position $q\in \mathbb{R}^{19}$ is 19 dimensional, and the first 7 elements are the floating-base joint (translation and rotation expressed via quaternion). 
The feature sets $\phi_e$ and $\phi_d$ are constructed in a similar way to that of the five-link robot example. 
We pick the task set $\Gamma$ to be walking with different speeds between $0.25$ and $0.75$ m/s and on different ground inclines between $-0.08$ and $0.08$ radians. 
For this example, to reduce runtime of the trajectory optimization subproblems, we chose a finite set of 9 tasks, evenly distributed in the task space.

Results, showing average cost per iteration, are shown in Fig. \ref{fig:cassie_2drom_cost} and \ref{fig:cassie_4drom_cost}. 
%We can see that there are less jittering because the tasks are fixed. 
As with the simpler example, reduced-order model optimization maintains model simplicity but dramatically improves performance.
Furthermore, we note that the final, optimized model, unlike its classical counterpart, does not map easily to a simple, physical model. 
While this limits our ability to attach physical meaning to $y$ and $\tau$, we believe this to be a necessary sacrifice to improve performance beyond that of hand-designed approaches.

\begin{figure}[t!]
 \centering
 \includegraphics[width=1\linewidth]{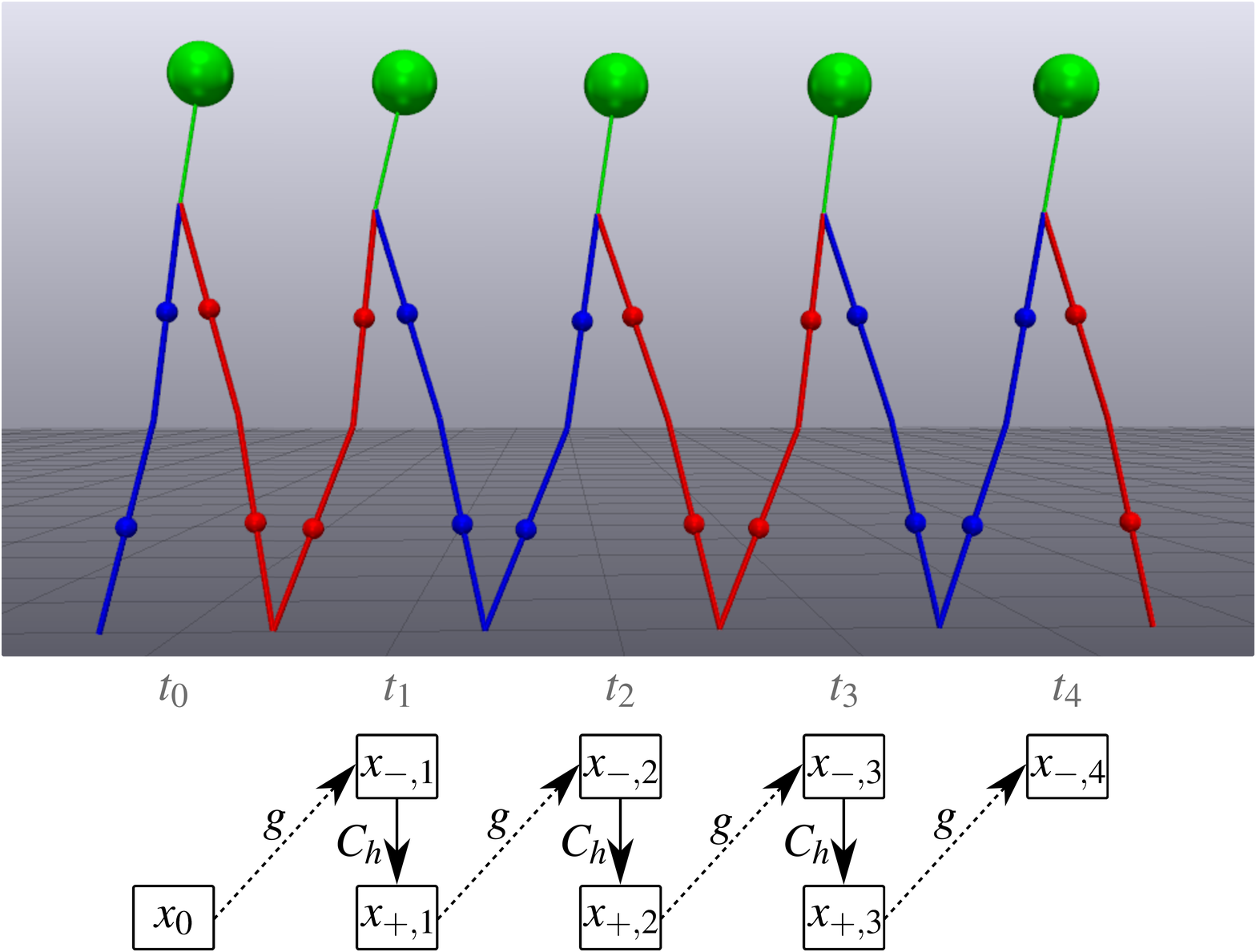}
 \caption{Given a task of covering two meters in five steps, we rapidly plan a trajectory for the reduced-order model. The high-dimensional model is used to capture the hybrid event, at stepping, as illustrated in the cartoon.}
 \label{fig:rom_planning}
\end{figure}

\section{Planning with Reduced-order Models}\label{sec:planning}

As shown in Fig. \ref{fig:outline}, given an optimal model $\mu^*$, we  plan in the reduced-order space.
As an example, we formulate a trajectory optimization problem to walk $l$ meters in $n_s$ strides.
Since the reduced-order model only captures the continuous dynamics, and perfect embedding of a reduced-order hybrid model is often impossible, we mix the reduced-order model with the discrete dynamics from the full-order model.
This approach results in a low-dimensional trajectory optimization problem, a search for $y_j(t)$ and $\tau_j(t)$, with additional decision variables $x_{-,j},x_{+,j}$, representing the pre- and post-impact full-order states.
The index $j=1,\ldots,n_s$ refers to the $j$th stride; note that the reduced-order model trajectories are necessarily hybrid.
Where $t_j$'s are the impact times (ending the $j$th stride), additional constraints relate these full-order states to the impact mapping and the reduced order model,
\begin{align*}
y_j(t_j) &= r(q_{-,j}; \theta_e), \quad &j=1,...,n_s-1,\\
y_{j+1}(t_j) &= r(q_{+,j}; \theta_e), \quad &j=1,...,n_s-1,\\
\dot{y}_j(t_j) &=\frac{\partial r (q_{-,j};\theta_e)}{\partial q_{-,j}} \dot{q}_{-,j},\quad  &j=1,...,n_s-1,\\
\dot{y}_{j+1}(t_j) &=\frac{\partial r (q_{+,j};\theta_e)}{\partial q_{+,j}} \dot{q}_{+,j},\quad  &j=1,...,n_s-1,\\
0&\geq C_{hybrid}(x_{-,j}, x_{+,j},\Lambda_j),\quad  &j=1,...,n_s-1.
\end{align*}
$C_{hybrid}$ represents the hybrid guard $S$ and the impact mapping $\Delta$ without left-right leg alternation.
Costs are nominally expressed in terms of $[y^\top, \dot{y}^\top]^\top$ and $\tau$, though the pre- and post-impact full-order states can also be used to represent goal locations.
This formulation preserves an exact representation of the hybrid dynamics, but results in a significantly reduced optimization problem that can be used for real-time planning. 

We tested the planning algorithm with the optimized 4D model embedded in the five-link robot. 
The distance $l$ varies from $0.2$ to $6$ meters with stride numbers $n_s$ between $1$ to $10$. 
To plan a single step, the average runtime was 110ms, on a computer with Intel i7-8750H processor, without optimizing code for efficiency.
Similar code required minutes for the full-order model.
Fig. \ref{fig:rom_planning} visualizes the pre-impact states in the case where the robot walks two meters with four strides, connected by the hybrid events and  continuous low-dimensional trajectories $y_j(t)$. 
We were able to retrieve $q(t)$ from $y_j(t)$ through inverse kinematics, meaning that the optimal trajectories $y_j(t)$ are feasible for the robot. 
The resulting motion, shown in the accompanying video, looks smooth and is qualitatively more efficient than the gait that the original model (LIP with a foot) would generate.

This demonstrates that the mixed model planner greatly reduces planning speed, and that the optimized reduced-order models can be used to achieve tasks in full-order space.

\section{Discussion}\label{sec:conclusion}
We present a novel method for automatically generating reduced-order models for legged locomotion, a step toward uncovering which aspects of the dynamics are necessary for tasks performance.
This approach is demonstrated over an array of tasks on both a simple, planar robot and a 3D model of the Cassie.
We also present an algorithm, suitable for real-time use, for planning reduced-order trajectories.

While there is no guarantee that the motions planned in Section~\ref{sec:planning} are feasible for the full model, we observe, empirically, that embeddings do seem to exist, and also note that classical models like LIP also provide no guarantees.
One direction for future work, for both LIP and the optimized models, is to generate constraints for reduced-order planning that guarantee feasibility on the original system.

Other future work will continue to develop and deploy reduced-order models, with an immediate goal of tracking and executing the planned motions on the physical Cassie robot using operational space control (e.g. \cite{Wensing2013}).
In Section \ref{sec:planning}, we note that the planner must still use the full-order model for the discrete mapping; future work will explore optimization  of hybrid reduced-order model.
Since impact maps are fully autonomous, it is not possible to find a perfect, low-order reduction.
This necessitates the need for approximate hybrid models, where we will leverage existing notions of hybrid distance \cite{Burden2015}.
Lastly, we would like to increase the tasks space and  explore alternative function bases to evaluate the quality of different resulting models.

%\section{FUTURE WORK}\label{sec:futurework}
%\input{future}


%%%%%%%%%%%%%%%%%%%%%%%%%%%%%%%%%%%%%%%%%%%%%%%%%%%%%%%%%%%%%%%%%%%%%%%%%%%%%%%%

%%%%%%%%%%%%%%%%%%%%%%%%%%%%%%%%%%%%%%%%%%%%%%%%%%%%%%%%%%%%%%%%%%%%%%%%%%%%%%%%

%%%%%%%%%%%%%%%%%%%%%%%%%%%%%%%%%%%%%%%%%%%%%%%%%%%%%%%%%%%%%%%%%%%%%%%%%%%%%%%%
%\section*{APPENDIX}

%\section*{ACKNOWLEDGMENT}

%%%%%%%%%%%%%%%%%%%%%%%%%%%%%%%%%%%%%%%%%%%%%%%%%%%%%%%%%%%%%%%%%%%%%%%%%%%%%%%%

\bibliographystyle{ieeetr}
\bibliography{library,yuming_library}

%\balance

\end{document}